\documentclass[10pt, conference]{IEEEtran}

\makeatletter
\usepackage[margin=0.75in]{geometry}
\usepackage{ifpdf}
\usepackage{cite}
\usepackage{amsmath}
\usepackage{algorithm}
\usepackage{multirow}
\usepackage[noend]{algpseudocode}
\ifCLASSINFOpdf
   \usepackage[pdftex]{graphicx}
 \else
\fi
\usepackage{upgreek}
\usepackage{dblfloatfix}
\usepackage{xcolor}
\usepackage{hyperref}
\usepackage{amssymb}
\usepackage{soul}
\usepackage{hhline}
\usepackage{booktabs}
\usepackage{float}
\hypersetup{
    colorlinks=true,
    linkcolor=blue,
    filecolor=magenta,      
    urlcolor=cyan,
    citecolor=cyan,
}
\IEEEoverridecommandlockouts

\title{\LARGE \bf 
 Before the Tipping Point: Force-Guided Active Perception for Shape-Agnostic Estimation of 3D Centers of Mass}

\begin{document}

\author{
    Steven~M.~Hyland,
    Jing Xiao,
    and~Cagdas~D.~Onal \\

    \thanks{The authors are affiliated with the Robotics Engineering Department, Worcester Polytechnic Institute, MA 01609, USA. All correspondence should be addressed to Steven Hyland \texttt{smhyland@wpi.edu}.}
} 

\maketitle

\begin{abstract}
Estimating the 3D center of mass (CoM) of unknown objects is challenging when grasping is infeasible, geometry is irregular, or mass distribution is uneven. We present a force-based method that estimates CoM height (i.e., $z_c$ value) and mass from a single sub-critical tipping experiment by a robot manipulator.
The robot applies a quasistatic elevated push and retract motion, using force–angle measurements recorded during tipping to identify parameters from the object trajectory. Our proposed push–retract cycle mitigates frictional bias, enabling generalized fitting. We experimentally validate our method using a robot manipulator with a six-axis force/torque sensor on varying types of objects without prior shape information and without specific models. We also propose a method to prevent toppling, keeping the object in a \emph{sub-critical} tipping regime by leveraging a safety margin. In experimental studies, our method recovers mass, CoM height, and toppling angle with relative errors below $\approx$5.0\% across all unknown objects. 
This work demonstrates reliable 3D inertial parameter estimation under proper safety thresholds in tipping. Our proposed method informs and enables reliable non-prehensile manipulation and robotic grasping of challenging objects that were previously infeasible.
\end{abstract}


\section{Introduction}
The Center of Mass (CoM) is a fundamental inertial property, playing a critical role in robotic manipulation—particularly in non-prehensile tasks. Accurate CoM estimation is necessary for stable grasping, efficient motion planning, and reliable prediction of object behavior under applied forces. While CoM can be computed for objects with known models of geometry and force distribution, estimating the CoM of previously unseen objects with unknown prior models remains challenging in unstructured settings. 

Non-prehensile manipulation offers a practical route to inference when grasping is infeasible due to object size, shape, fragility, or task constraints. We introduce an approach where a robot executes controlled pushes on an unknown object and collects sensory data to infer the CoM height and mass. This process forms an active perception loop, where purposeful interaction with the object and environment incrementally improves the robot’s understanding of the scene.

We leverage repeatable quasistatic interaction regimes with complementary sensing: visual sensing provides kinematic context (such as the pose of the interacted object), while force measurements expose the physical consistency of tipping behavior. Building on an estimate of the planar 2D $(x_c\,,y_c)$ CoM projection, we use controlled tip-inducing pushes to resolve CoM height and recover mass and critical toppling angle without requiring destructive toppling. Toppling is a dangerous condition in which the object uncontrollably falls over, which we avoid by implementing a \emph{safety margin} ($\eta_\text{safety}$) and admissible force envelope ($\mathcal{F}$) to constrain the system's behavior.

\section{Prior Work}

Estimating inertial properties of unknown objects through interaction is a long-standing topic in robotics, motivated by the fact that mass and CoM govern object motion and stability under applied wrenches. Much of the prior work emphasizes 2D parameter estimation (i.e. planar components on the supporting surface). Earlier work focused on the effects of different support surfaces, such as wood, ceramic, and stainless steel on pushing operations and estimates \cite{tanaka_active_2004}. More recent efforts focused on using tactile feedback \cite{yuan_active_2023}, using combined grasp-push strategies \cite{uemura_dynamic_2024}, and data-driven approaches \cite{mavrakis_estimating_2020}.



\subsection{Grasp-Based Methods}
There is a large body of work that estimates inertial parameters under the controlled conditions afforded by grasped or end-effector-fixed objects. In this setting, the robot excites the fixed object while measuring wrist force/torque and end-effector motion, enabling classical identification of CoM, mass, and inertia. Atkeson {\em et al.} and {\em Khosla et al.}  each approached this system identification problem using excitation trajectories, combined with least-squares estimation of linearized Newton-Euler equations \cite{atkeson_rigid_1985, khosla_parameter_1985}. Typical motions included lifting, shaking, or other pre-planned motions, until Gautier and Swevers {\em et al.} optimized trajectories to improve accuracy and reduce uncertainty \cite{gautier_dynamic_1997, swevers_optimal_1997, swevers_dynamic_2007}. Later extensions focus on cooperative multi-robot identification that distributes the applied wrench across agents, both with mobile and aerial robot platforms \cite{marino_distributed_2017, corah_active_2017, mas_object_2012}. While highly accurate, these approaches depend on stable grasping and payload constraints that are often unavailable in non-prehensile scenarios.

\subsection{Planar Pushing and Planar CoM}
Planar pushing has been extensively studied both as a mechanism for moving objects on a support surface and as a rich source of information for inferring physical parameters. Early work by Mason {\em et al.} established the mechanics of pushing and introduced the \emph{voting theorem}, relating observed object rotation to the center of friction and friction cone \cite{mason_mechanics_1986}. 

Building on this foundation, several works began employing planar pushing for \emph{active inference}. Rather than merely predicting motion, these approaches use selective pushing to elicit informative behaviors from the object - a strategy with strong ties to grasp-based methods. Developments in this area span open- and closed-loop pushing control, stable fence pushing, and initial techniques for inertial parameter estimation \cite{yoshikawa_indentification_1991, lynch_manipulation_1992, lynch_estimating_1993, lynch_dynamic_1996, akella_posing_1998}. Further advancements in push-manipulation include methods based on rapidly exploring random trees (RRTs) for graspless manipulation and dual-finger pushing \cite{miyazawa_planning_2005, yong_yu_estimation_2005}. While studies departing from quasistatic regimes are limited, impact-based interactions have also been explored to infer density distribution \cite{mkhitaryan_visual_2013}.

Despite its convenience, planar interaction alone cannot resolve the CoM height, as distinct 3D CoM configurations can share identical planar projections. This fundamental ambiguity necessitates out-of-plane motions such as tipping. In this work, we assume an estimate of the planar CoM projection is available (e.g., from prior planar pushing methods such as \cite{hyland_predicting_2023, hyland_onboard_2025}) and focus on the vertical component. 

\subsection{Tipping and Toppling for CoM Inference}
Object tipping has long been used in mechanics to characterize a system's stability margin. In non-prehensile manipulation,  this physical reasoning underpins tasks like safe pushing, toppling prevention, and stability-aware planning. 

Despite its utility, this area remains relatively under-studied compared to planar pushing. A seminal contribution to 3D inference was provided by Yong Yu {\em et al.}, who introduced the \emph{Gravity Equi-Effect Plane} (GEEP), defined as a plane containing the CoM and the object's pivot line-contact with the support surface \cite{yong_yu_estimation_1999}. When the GEEP aligns with the gravity vector, the object is \emph{marginally stable}; by tipping or tilting the object about multiple contact lines, the 3D CoM can be resolved from the intersection of the resulting planes. This framework was later extended to cylinder-like objects and curved bases where a ``passing-CoM line" is used in place of the equi-effect plane to account for moving point-contacts at the base \cite{yu_estimation_2001}. However, the approach requires considerable prior information of the object geometry. 



\section{Approach}
Our approach aims to estimate an object's CoM without knowing the object geometry, without requiring that the object is graspable, and without using multiple tipping operations. We use a two-stage pushing approach:
\begin{enumerate}
    \item We first conduct 2D CoM estimation: planar coordinates $(x_c,y_c)$ of CoM are determined by applying pushes at low height, which induces in-plane motion. 
    
    \item Next, we select a higher pushing point to induce tipping, allowing for the estimation of the CoM height $z_c$ and object mass $m$. We emphasize sub-critical tipping for enhanced safety and repeatability without destructive toppling, and recover mass and critical toppling angle as natural byproducts. This single-tipping inference eliminates the need for multiple push/tip configurations.
\end{enumerate}

We rely on fundamental principles of mechanics and integrate force and visual sensing to monitor applied forces and object poses and motion. The analysis is conducted under the following assumptions:
\begin{enumerate}
    \item The manipulated object is a non-deformable rigid body.
    \item Isotropic Coulomb friction exists. We use the simplification of a single friction coefficient, $\mu$, for both static and kinetic friction.
    \item Quasistatic motion is conducted such that inertial influences are negligible.
\end{enumerate}




\begin{figure*} [!tp]
    \centering
    \includegraphics[width=\linewidth]{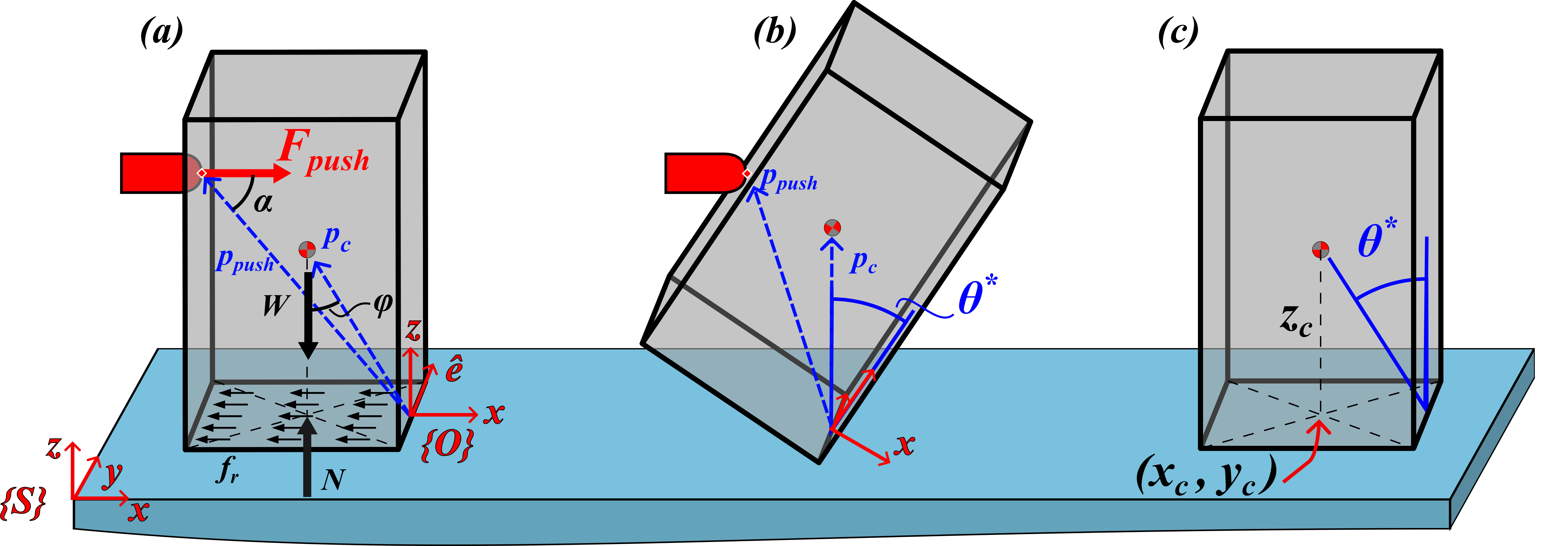}
    \caption{Object being tipped by a finger applying external force $\mathbf{F}_\text{push}$ in its (a) resting equilibrium phase, its (b) toppling equilibrium phase with zero pushing force, and (c) the characteristic $\theta^*$ associated with the object. $\theta^*$ is found by visual sensing of the object's pose at the toppling threshold. \textbf{Important:} All vectors are defined and observed with respect to the object frame $\{O\}$, which is located along the pivot edge. $\alpha$ is the angle between applied force and pivot vectors, while $\varphi$ is the angle between the gravity and pivot vectors.}
    \label{fig: boxes}
\end{figure*}




\subsection{2D CoM Estimation}
Low-height pushing minimizes the likelihood of tipping and by observing the object's rotational behavior, the 2D CoM coordinates may be extracted. In \cite{hyland_predicting_2023}, this is performed by iteratively selecting a new push direction at constant low-height that minimizes object rotation. This way, the line of action that passes through the 2D CoM projection can be found. Performing a second push from another face of the object yields a second line of action; the intersection of both lines defines the 2D CoM. In the present study, we assume that this planar CoM information is already extracted.

\subsection{Sliding Condition}

For pushing at a high height, such as shown in Figure \ref{fig: boxes}\textcolor{blue}{(a)}, both sliding and tipping must be considered. The sliding condition can be found by analyzing the force balance of the rigid body system. All frictional forces have an upper limit determined by the normal force magnitude $N$ and the static friction coefficient $\mu_s$ \cite{lynch_modern_2017}. Incipient sliding occurs when the force magnitude applied by the finger reaches this maximum static friction threshold. 


\subsection{Tipping Condition}
Pure tipping is defined by the instant where any vertex of the support polygon breaks contact with the support surface. Normal quasistatic tipping always occurs about the surface contact(s) that are furthest from the pushed side. The instantaneous axis of rotation defining the contact(s) are henceforth referred to as the body's \emph{pivot}.

We can define the torque-balance equation generalized for any rigid body tipped by an arbitrary external force:
\begin{equation} \label{eq: f_tip}
    \boldsymbol{\tau}_\text{push} + \boldsymbol{\tau}_\text{grav} = 
    (\mathbf{p}_\text{push} \times \mathbf{F}_\text{push})
    +
    (\mathbf{p}_c \times \mathbf{W}) = 0 \;\;\; \in \mathbb{R}^3
\end{equation}

\noindent We further define the following symbols, as illustrated in Figure \ref{fig: boxes}: 
\begin{itemize}
    \item $\{S\}$ world coordinate frame and $\{O\}$ object frame.
    \item $\hat{\mathbf{e}} \in \mathbb{R}^3$ unit vector defining the tipping axis.
    \item $\theta \in [0, \pi)$ object angle about $\hat{\mathbf{e}}$ in world coordinate frame.
    \item $\mathbf{p}^S_O$ object frame position in the world coordinate frame.
    \item $\mathbf{W}^O(\theta) = R(\text{-}\theta)\mathbf{W}^S$ object weight in object frame.
    \item $\mathbf{p}^O_\text{push}(\theta) = \mathbf{p}_\text{push}^S - \mathbf{p}_O^S$ fingertip position in object frame.
    \item $R(\theta)$ rotation matrix associated with object rotation $\theta$.
    \item $\mathbf{p}_{c(x,y)}$ projected planar CoM position in object frame.
    \item $\varphi = \angle(\mathbf{p}_c, \mathbf{W})$ angle between CoM and weight vector.
    \item[]
\end{itemize}
\noindent These definitions can be substituted into \eqref{eq: f_tip}:
\begin{equation} \label{eq: generalized_final}
    \boxed{
        \mathbf{p}_\text{push}^O \times \mathbf{F}^O
        +
        \mathbf{p}^O_c \times \mathbf{W}^O = 0 \quad \in \mathbb{R}^3
    }
\end{equation}

Equation \eqref{eq: generalized_final} serves as the model upon which the unknown parameters are estimated. Since the majority of the following analysis is in the object local frame, unless otherwise notated, all subsequent terms are with respect to $\{O\}$.

\subsection{Mode Selection at Onset}
During monotonic pushing, the first mode to appear is the one whose balance equation is first satisfied. This is due to the resistive sliding friction or gravity moment remaining constant, while the applied force $\mathbf{F}_\text{push}$ gradually increases.

\noindent\textbf{Sliding occurs when:}
$$
    \|\mathbf F_\text{push}\|\;\ge\;F_{\text{slide}} 
    =
    \mu_s N \, , \quad \text{where } N\text{ is normal force}.
$$
\noindent\textbf{Tipping occurs when:}
$$
    \|\mathbf F_\text{push}\|\;\ge\;F_{\text{tip}}
    =
    \frac{\|\mathbf{p}_c\times\mathbf{W}\|}{\|\mathbf{p}_\text{push}\|\sin\alpha}, 
    \quad \text{where } \alpha \text{ = } \angle(p_\text{push}, F).
$$

The inequality $F_{\text{slide}} \stackrel{?}{\gtrless} F_{\text{tip}}$ determines which mode will occur first. When both are equal, a coupled slide–tip onset occurs. While sliding force ($F_{\text{slide}}$) can be constant per object-surface couple based on geometry, weight, and friction, tipping force ($F_{tip}$) is a variable dictated by the push height and push direction. 

\noindent\textbf{Toppling} is the instant that the body's CoM is directly above the body's pivot, when the CoM's projection onto the support surface coincides with the pivot. At this body angle, $\theta^*$, the system is in marginally stable equilibrium with zero pushing force, and any infinitesimal perturbation causes a fall. 

\noindent\textbf{Backwards Tipping} is only possible during dynamic pushing, if an impulse is imparted \emph{below} the CoM. The quasistatic pushing motion eliminates this possibility. 

\noindent\textbf{Feasibility:} 
As $\mathbf{F}_\text{push}$ is directed towards the pivot (i.e., $\alpha \Rightarrow 0$), the applied moment falls to zero and thus tipping cannot occur. Conversely, as $\mathbf{F}_\text{push}$ becomes orthogonal to the pivot (i.e., $\alpha=90^\circ$), tipping leverage is maximized. For certain objects, especially those with low-height CoM and wide bases, $F_\text{tip}$ may be impossibly large under Coulomb friction and achieving it may demand a tangential (vertical) component exceeding contact friction limits at the fingertip. We treat such objects as outside the admissible class of the present tipping method. 

\begin{figure*} [!tp]
    \centering
    \includegraphics[width=\linewidth]{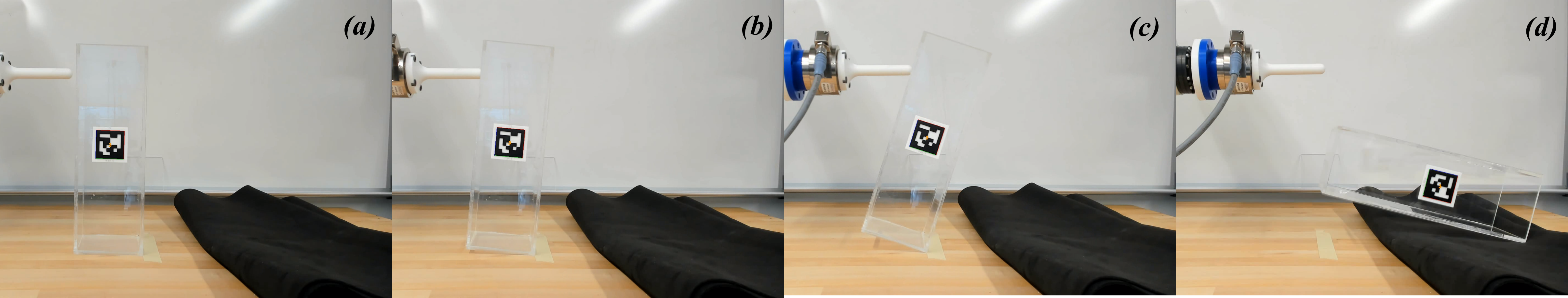}
    \caption{Full toppling procedure using the box object, which is tipped through the toppling threshold, at which point the object topples over onto the table surface. Our shape-independent force-based approach successfully infers parameters despite the object's clear appearance, which makes shape detection difficult by vision.}
    \label{fig: exp_topple}
\end{figure*}

\subsection{Estimation of CoM z-coordinate}
The tipping torque balance elucidates many characteristics of the relationship between the applied external force and the object's tipping angle. Observe that the gravity torque magnitude in \eqref{eq: generalized_final} decreases as the weight vector and CoM position vector approach collinearity (i.e., as $\varphi \Rightarrow 0$):
$$
    ||\mathbf{p}_c \times R(\text{-}\theta)\mathbf{W}^S|| = ||p_c|| \cdot||W^S|| \sin(\varphi) 
$$
\noindent Note: $\varphi$ is a purely object-specific parameter influenced only by the object angle $\theta$. 

As a result, to maintain quasistatic torque balance, with $\mathbf{p}_\text{push} \ne 0$, the magnitude of the fingertip force, $||\mathbf{F}_\text{push}||$, must in turn decrease:
\begin{equation} \label{eq: limit}
    \lim_{\varphi \rightarrow 0} ||\boldsymbol{\tau}_\text{grav}|| 
    =
    \lim_{\theta \rightarrow \theta^*} ||\mathbf{F}_\text{push}||
    =
    \begin{bmatrix}
        0 & 0 & 0 
    \end{bmatrix}^\top
\end{equation}

The payload's angle at this toppling threshold is referred to as the \textbf{toppling angle}, $\theta^*$. Given that variables $\mathbf{F}_\text{push}$, $\mathbf{p}_\text{push}$, and $\theta$ are all measurable, only unknown constants remain: CoM height and mass ($z_c^O, m$). Equation \eqref{eq: generalized_final} provides two methods of estimating these parameters. 
\newline 


\noindent \textbf{1. Geometric computation:} Observing Figure \ref{fig: boxes}\textcolor{blue}{(c)}, the toppling angle $\theta^*$ can be experimentally determined by leveraging the relationship in \eqref{eq: limit} via measurement of the toppling angle at which gravity torque drops to zero. The CoM height can then be found by geometric analysis:
\begin{equation} \label{eq: geom_zc}
    z_c = \frac{\sqrt{x_c^2 + y_c^2}}{\tan{\theta^*}}
\end{equation}
\noindent and the mass can be determined by solving the torque balance for $\mathbf{W}^S=[0, 0, -mg]^\top$  analytically or through numerical methods. \newline

\noindent \textbf{2. Physics-based curve fit:} By collecting various ($\mathbf{F}_\text{push} \,, \theta$) coupled datapoints, we can use regression on our model in \eqref{eq: generalized_final} to extract the unknown parameters, $z_c$ and $m$, that best match the measured data. Our subsequent experiments aim at determining these values and make use of both geometric and curve fit methods. 
%

\subsection{General Procedure}
The robot executes a controlled push experiment at a low constant velocity to induce quasistatic tipping of the object. 
Each trial consists of a single quasistatic push–retract interaction from which the inertial parameters are estimated: 

\begin{enumerate}
    \item Use the planar CoM projection $(x_c,y_c)$ to select a pushing location at the top of the object that maximizes tipping leverage while avoiding 
    sliding.
    \item Apply a quasistatic horizontal push and retract cycle to the object at a constant height.
    \item Record the coupled force–angle trajectory $(\mathbf{F}_\text{push},\theta)$ throughout the tipping motion.
    \item Perform signal conditioning on force trajectories using a cascaded median and Savitzky--Golay filter to suppress high-frequency noise while preserving transient tipping characteristics. 
    \item Perform a first-order linear regression on the filtered dataset to derive a heuristic estimate of $(z_c,\text m)$, which serves as a \emph{warm start} for the subsequent optimization. Outliers are rejected using median absolute deviation (MAD) from the linear fit.
    \item Refine the parameters using model-based nonlinear least-squares (NLS) applied to the coupled $(F,\theta)$ manifold to extract the underlying physical parameters.
\end{enumerate}

Although contact friction is not explicitly modeled in \eqref{eq: generalized_final}, its influence is reduced through a single push–retract cycle. The reversal of motion induces a corresponding reversal in frictional direction, producing a characteristic hysteresis in the measured force–angle trajectory. By combining both the push and retract segments for regression to the model, friction-induced bias is effectively canceled, yielding parameter estimates robust to unmodeled frictional effects.

\subsection{Sub-Critical Single-Push Estimation}

Observing Figure \ref{fig: exp_topple}, reaching the toppling threshold is undesirable in practical manipulation settings, as $\theta^*$ corresponds to a state of marginal stability. Operating near this condition increases the risk of uncontrolled toppling and potential damage to the object or environment. To ensure safe interaction, we instead enforce a sub-critical tipping policy in which the object is intentionally kept below the toppling angle, i.e., $\theta < \theta^*$.

In practice, $\theta^*$ is a latent parameter unknown a priori. It would be impractical to first topple an object to identify its toppling angle only to avoid that state in subsequent interactions. Instead, we exploit the monotonic force-angle coupling: within the quasistatic regime, the force magnitude provides a reliable proxy for proximity to $\theta^*$. We regulate the interaction by constraining the measured force, $\mathbf{F}_\text{push}$, to lie within a bounded \emph{admissible envelope} $\mathcal{F}_\text{adm}$:
$$
    \mathbf{F}_\text{push}(\theta)\in \mathcal{F}_\text{adm}, \;\; 
    \text{where }
    \mathcal{F}_\text{adm} = \{\mathbf{F}_\text{push}|F_\text{safe} \leq F_\text{push} \leq F_\text{max}\}.
$$ 

Here, $F_\text{max}$ denotes the peak interaction force magnitude observed at initial contact, while $F_\text{safe}$ is a tunable safety floor. The value of $F_\text{safe}$ is selected according to application-specific safety requirements.

To study the effect of this constraint, we parameterize the safety threshold using the dimensionless safety margin $\eta_\text{safety}$:
$$
    F_\text{safe} = \eta_{\text{safety}} F_{\text{max}}, 
    \quad 
    \eta_\text{safety} \in (0,1).
$$

This formulation enables geometry-agnostic comparison of estimation performance across objects and safety margins.


\section{Experiments, Results,  and Discussion}

Experiments were conducted using an ABB IRB120 industrial manipulator equipped with a rigid 3D-printed fingertip that provides point contact with the object. The robot primarily pushes in the $\text{+}x$ direction while interaction forces are measured using a wrist-mounted force/torque sensor. Visual feedback is obtained utilizing a fixed web camera, which observes the object pose from an AprilTag affixed to the object. This configuration provides synchronized measurements of the applied wrench and object response required for parameter inference. All model fitting was performed on an Intel Core i7-13700H 14-core processor.

We used a reference object to validate the efficacy of our approach. It is a hollow acrylic rectangular prism with its CoM located at the geometric centroid. This object is henceforth referred to as the \emph{Box object}.

\begin{figure} [!t]
    \centering
    \includegraphics[width=\linewidth]{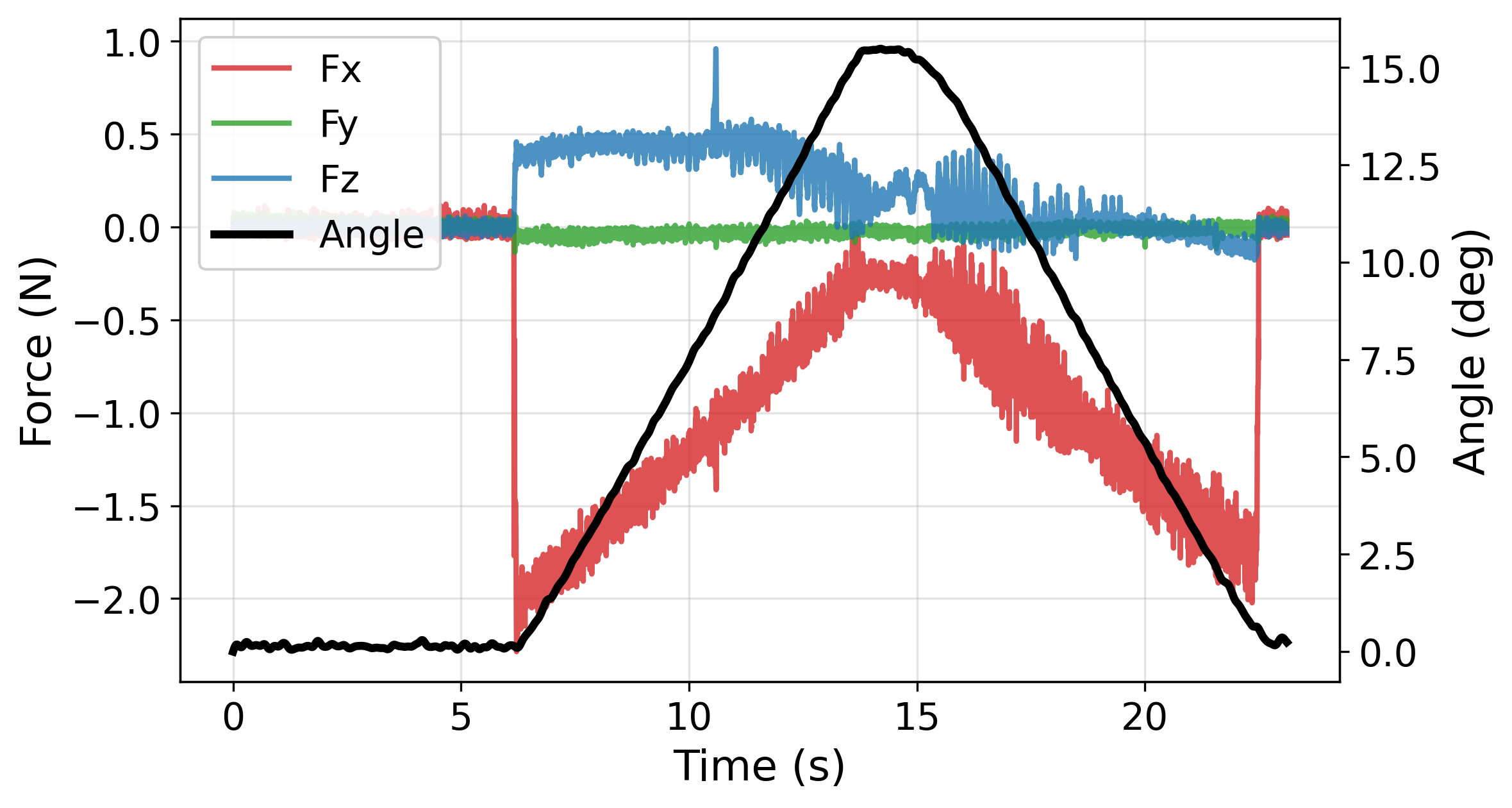}
    \caption{Experiment: Live raw data stream for a representative sub-critical tipping trial ($\eta_\text{safety}=0.1$) of the reference box object. Measured hysteresis between phases captures frictional direction reversal, while the object angle changes during quasistatic tipping. Applied torque is calculated from measured force and finger position. A small spike in force is noted around ten seconds due to imperfect box surface characteristics.}
    \label{fig: box_0.1_live}
\end{figure}

\begin{figure} [!h]
    \centering
    \includegraphics[width=\linewidth]{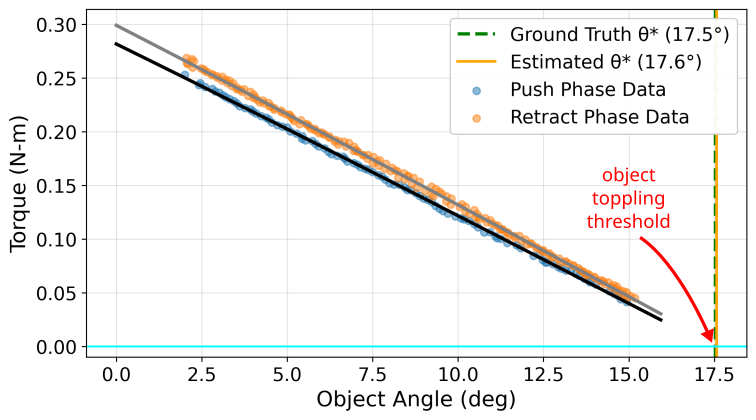}
    \caption{Experiment: Torque-angle model fit for the box object at $\eta_\text{safety}=0.1$. Changing friction direction causes push/retract phase offset. The fitted parameters $(z_c,m)$ align with the expected force decay as $\theta\rightarrow\theta^*$.}
    \label{fig: box_0.1_fit}
\end{figure}

\subsection{Box Reference Object Sub-Critical Tipping}
We first evaluated estimation accuracy under sub-critical tipping using the reference box object. A liberal safety margin of $\eta_\text{safety}=0.1$ was selected to approximate the conditions of the full-toppling validation while ensuring that the interaction remained strictly below the toppling threshold. Figure \ref{fig: box_0.1_live} shows the raw measured force and angle time profiles.

\begin{table}[!t]
  \centering
  \caption{Experimental estimate errors for box ($\eta_{\mathrm{safety}}=0.1$).}
  \label{tab: box_exp}
  \begin{tabular}{lccc}
    \toprule
     & $m$ (kg) & $z_c$ (cm) & $\theta^*$ (deg)\\
    \midrule
    \textbf{Ground truth}  & \textbf{0.658} & \textbf{14.624} & \textbf{17.532}\\
    \addlinespace[2pt]
    Push phase             & 0.622          & 14.71          & 17.435\\
    Retract phase          & 0.660          & 14.48          & 17.700 \\
    \addlinespace[2pt]
    \textbf{Mean} & \textbf{0.641} & \textbf{14.594} & \textbf{17.566} \\
    \midrule
    Absolute error         & -0.017         & -0.030          & 0.034\\
    Relative error         & 2.618\%        & 0.205\%         & 0.195\% \\
    \bottomrule
  \end{tabular}
\end{table}

Figure \ref{fig: box_0.1_fit} demonstrates the corresponding fit and reveals a clear separation between the push and retract segments of the trajectory, arising from direction-dependent friction at the finger–object contact. Table~\ref{tab: box_exp} lists the sub-critical tipping errors, which are low across all estimated parameters. The push and retract segments individually bracket the ground-truth values due to frictional bias, while their combined estimate provides the most accurate result. This behavior is consistent with the expected hysteresis structure of the interaction and confirms that reliable CoM height and mass estimates can be recovered without approaching toppling.

Having established accurate performance on the reference object under sub-critical conditions, we next evaluate the method on previously unseen objects with varying geometries and contact properties.

\subsection{Application to Unknown Objects of Different Geometries}
\begin{figure} [!t]
    \centering
    \includegraphics[width=\linewidth]{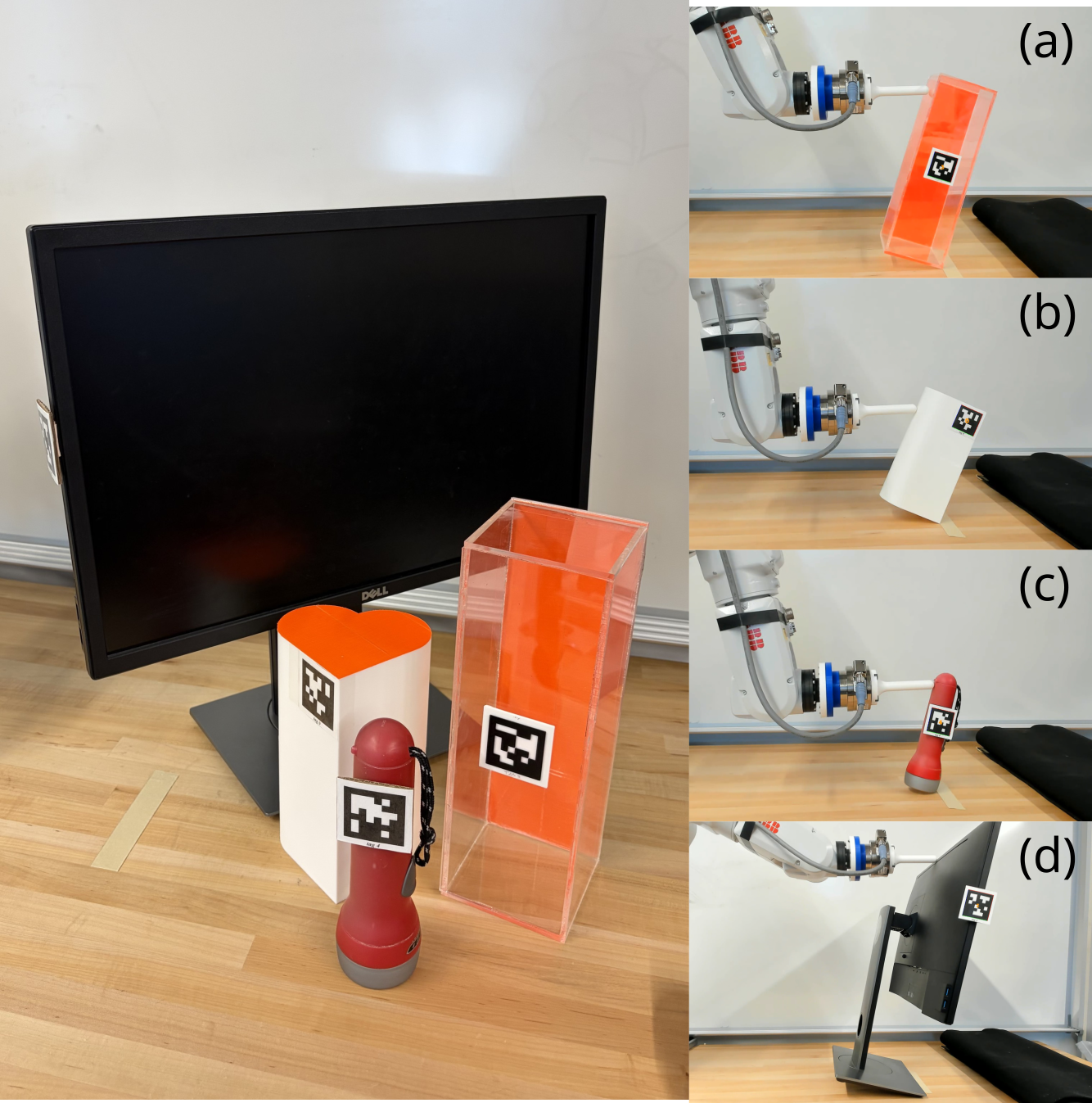}
    \caption{Objects used for experiments. Box (a) and heart (b) objects are designed and 3D printed. Flashlight (c) and monitor (d) are everyday objects. The rear face of the box was colored orange to improve contrast for the reader; the approach is unaffected by this visual change.}
    \label{fig: all_objects}
\end{figure}

Figure \ref{fig: all_objects} shows the candidate objects on which we evaluate the proposed method, including the reference box object. The remaining three objects are a \emph{heart-shaped prism}, a \emph{handheld flashlight}, and a \emph{computer monitor}. All four candidate objects each exhibit distinct geometries, frictional characteristics, mass distributions, and contact profiles.

\begin{table}[!t]
  \centering
  \caption{Ground truths for each object.}
  \label{tab: ground_truths}
  \begin{tabular}{lccc}
    \toprule
     & $m$ (kg) & $z_c$ (cm) & $\theta^*$ (deg)\\
    \midrule
    \textbf{Box}  & \textbf{0.658} & \textbf{14.624} & \textbf{17.532}\\
    \midrule
    \textbf{Heart} & \textbf{0.219} & \textbf{9.800} & \textbf{23.984} \\
    \midrule
    \textbf{Flashlight} & \textbf{0.336} & \textbf{9.656} & \textbf{15.126} \\
    \midrule
    \textbf{Monitor} & \textbf{5.040} & \textbf{23.201} & \textbf{14.748} \\
    \bottomrule
  \end{tabular}
\end{table}

\begin{figure*} [!h]
    \centering    \includegraphics[width=\linewidth]{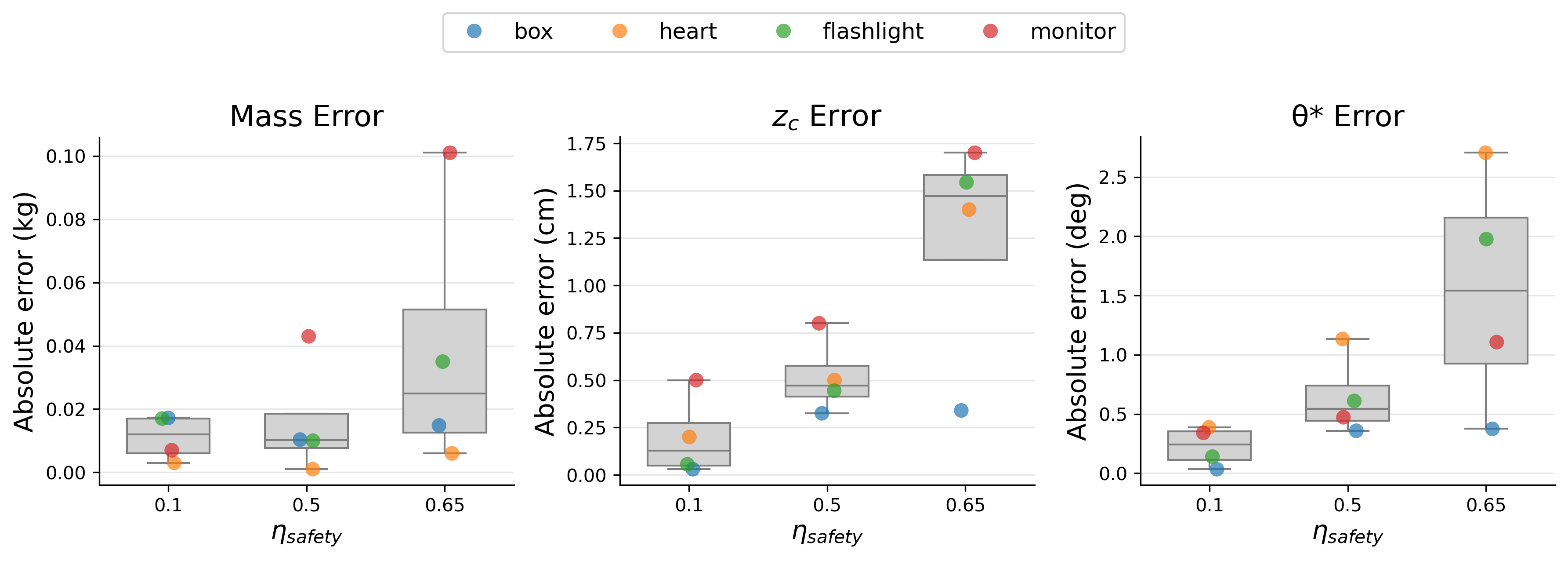}
    \caption{Absolute error in mass, CoM height, and toppling angle for all tested objects as a function of the safety margin parameter $\eta_\text{safety}$. Each box plot aggregates results from each of the four objects for a particular safety margin. Errors increase with conservative safety settings due to reduced observability near the toppling threshold, though performance remains within acceptable bounds for practical manipulation.}
    \label{fig: abs_error_summary}
\end{figure*}

\begin{figure*} [!h]
    \centering    \includegraphics[width=\linewidth]{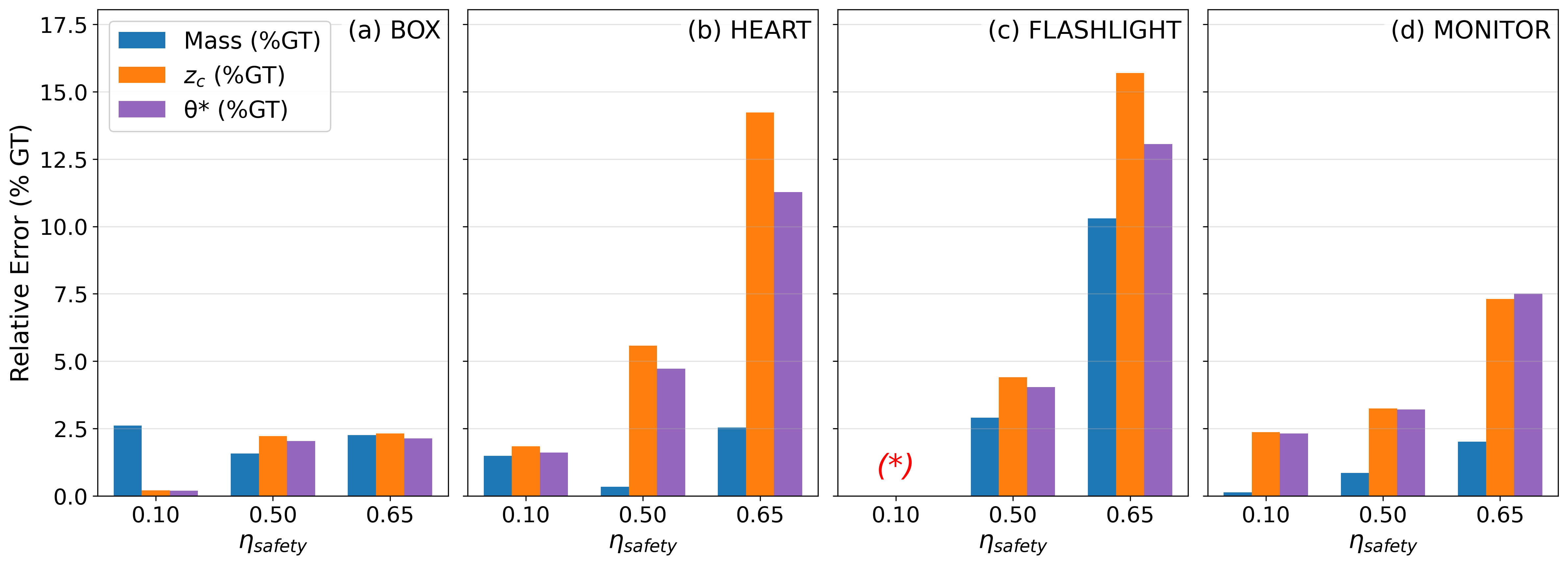}
    \caption{Relative error in mass, CoM height, and toppling angle for each object at varying safety margins. Mass estimates remain consistently stable across settings, while CoM height and toppling angle exhibit increased sensitivity to truncated trajectories. (*) The flashlight object at $\eta_\text{safety}=0.1$ lacks a valid estimate due to instability induced by its curved base, illustrating a failure mode of tipping-based inference.}
    \label{fig: perf_summary}
\end{figure*}

We first determined the ground truth dimensions and parameter values of these objects, as shown in Table \ref{tab: ground_truths}. The mass was measured with a weighing scale, while $z_c$ was determined by leveraging the technique described in \cite{mcgovern_learning_2019}, wherein the object was placed at the edge of the support surface until tipping was first detected. Then, the distance from the object frame $\{O\}$ to the surface edge was measured for the given CoM axis. This was repeated for all three axes to calculate the ground truth CoM.

Figure \ref{fig: abs_error_summary} summarizes the aggregate absolute errors. The single push–retract estimation procedure was applied to each object without modification besides push height selection. The individual raw trajectories and fit plots are omitted for brevity. Overall estimation performance is shown in Figure~\ref{fig: perf_summary}.

\begin{table}[!t]
\centering
\caption{Fit time and push--retract completion time per object and safety margin.}
\label{tab: times}
\begin{tabular}{llcc}
\toprule
Object & $\eta_\text{safety}$ & Fit (ms) & Cycle (s) \\
\midrule

\multirow{3}{*}{Box}
  & 0.1  & 12.08 & 22.583 \\
  & 0.5  & 6.91 & 12.739 \\
  & 0.65 & 5.83 & 8.829 \\

\midrule
\multirow{3}{*}{Heart}
  & 0.1  & 11.74 & 19.225 \\
  & 0.5  & 6.84 & 6.632 \\
  & 0.65 & 4.90 & 6.434 \\

\midrule
\multirow{3}{*}{Flashlight}
  & 0.1  & \text{-} & \text{-} \\
  & 0.5  & 5.14 & 7.084 \\
  & 0.65 & 4.81 & 5.217 \\

\midrule
\multirow{3}{*}{Monitor}
  & 0.1  & 14.51 & 31.566 \\
  & 0.5  & 9.06 & 19.931 \\
  & 0.65 & 8.20 & 15.624 \\

\bottomrule
\end{tabular}
\end{table}

Table \ref{tab: times} lists the estimation convergence time and the push-retract cycle time for each object-safety-margin pair. Although we do not optimize for convergence time in the present work, and we utilize identical push speeds across all objects, there are methods to improve these rates, which we discuss in Section~\ref{sec:conclusions}. We would expect there to be a positive correlation between cycle time and $\theta^*$ since the object would seemingly reach its toppling threshold sooner at lower values of $\theta^*$. \emph{Instead}, we observe a positive correlation between \emph{cycle time} and \emph{object mass}; the lighter objects had quicker push–retract cycles.

This behavior is explained by the safety-limited admissible envelope. The push phase is terminated when the applied force reaches a fraction $F_\text{safe}\text{ = }\eta_\text{safety}F_\text{max}, \text{ where } F_\text{max}$ corresponds to the force at incipient tipping. Since the resistive torque scales with the object weight, heavier objects require larger applied forces before the safety condition is met. Under identical push velocities across objects, this results in longer force build-up durations and therefore longer push–retract cycles. Consequently, cycle time is governed primarily by mass-dependent force thresholds rather than by the tipping angle $\theta^*$ itself.

Two consistent trends emerge across objects and safety margins. First, mass estimates remain comparatively stable across the tested range of $\eta_\text{safety}$, while $z_c$ and $\theta^*$ exhibit greater sensitivity to truncation and measurement noise. This behavior is expected: mass recovery depends primarily on the magnitude and slope of the measured wrench trajectory (as reflected in $\mathbf{W}(\theta)$ in \eqref{eq: generalized_final}), which is observable throughout the interaction. In contrast, $z_c$ depends more strongly on accurate inference of $\theta^*$ and is therefore more sensitive to early truncation of the trajectory. Additionally, across all but one experiment (monitor $\eta_\text{safety}=0.65$), $\theta^*$ estimates exhibit lower error than $z_c$. This is caused by the fact that our coupled force-angle data is actually directly measuring the $\theta^*$ value, whereas the parameter estimation is extrapolated from this data. Therefore errors are amplified during the indirect estimation of $z_c$.

Second, the safety margin $\eta_\text{safety}$ induces a clear tradeoff between safety and accuracy. Conservative settings (larger $\eta_\text{safety}$) terminate the interaction earlier, reducing risk but limiting the extent to which the data capture the characteristic force decay as $\theta \rightarrow \theta^*$. Under these conditions, $\theta^*$ must be inferred primarily through extrapolation, which can amplify error in $z_c$. More aggressive settings (smaller $\eta_\text{safety}$) allow deeper tipping and improve parameter estimation under ideal conditions, but also increase exposure to non-ideal behaviors such as pivot-edge transitions, intermittent slip, or local compliance that violate quasistatic assumptions. This tradeoff is reflected in the object-dependent performance spread observed in Figure~\ref{fig: perf_summary}.

The flashlight and heart are the lightest objects tested, producing comparatively small interaction wrenches and therefore lower signal-to-noise ratios. This sensitivity is exacerbated at high safety margins, where the usable trajectory is truncated further.

Across all tested objects, no single $\eta_\text{safety}$ value is universally optimal. While multi-push strategies could be used to adaptively select a safety level by comparing estimates obtained at different margins, such a procedure would reduce efficiency and may be impractical in many applications. In practice, selecting $\eta_\text{safety}$ according to object class or task risk tolerance provides a more realistic operating strategy. For these experiments, values $\eta_\text{safety}<0.65$ produced usable trajectories, while larger values truncated interaction too early to support reliable fitting.

\begin{figure} [!t]
    \centering
    \includegraphics[width=\linewidth]{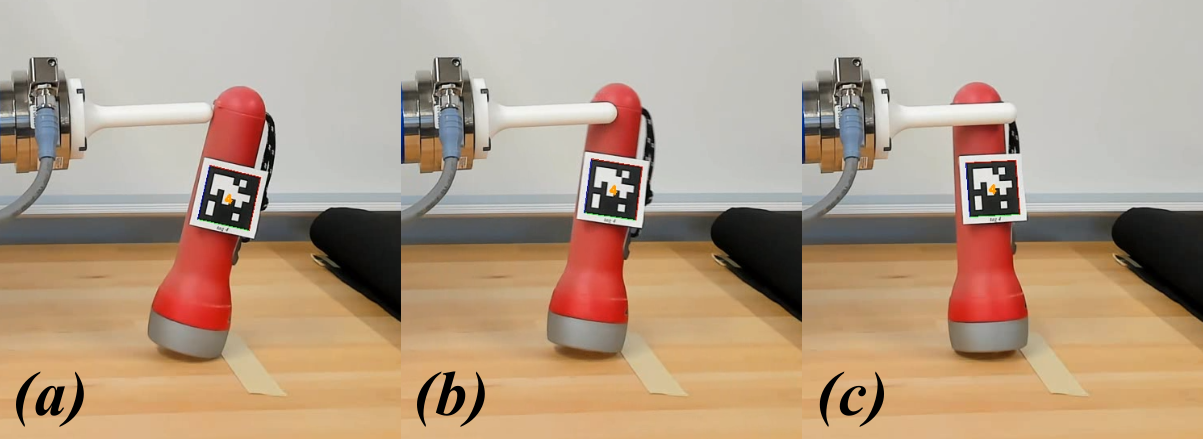}
    \caption{Flashlight object failure mode: low safety margin ($\eta_\text{safety}=0.1$) causes flashlight to swirl around its circular base, despite the finger line of action appearing to pass through the planar $(x_c,y_c)$ CoM. Finger moves towards the right with sequence (a), (b), (c).}
    \label{fig: flashlight_fail}
\end{figure}

Figure \ref{fig: flashlight_fail} demonstrates a failure mode encountered when tipping an object with a curved base. The flashlight object at $\eta_\text{safety}=0.1$ is particularly prone to instability, and does not yield a valid estimate. Its round edge produces shifting contact geometry during tipping, introducing yaw motion 
and instability near the toppling region, violating the fixed-pivot assumption of the model. This failure mode illustrates a limitation of single-contact forward tipping for curved-base geometries and motivates interaction strategies that more directly stabilize the pivot.

\begin{figure} [!b]
    \centering
    \includegraphics[width=\linewidth]{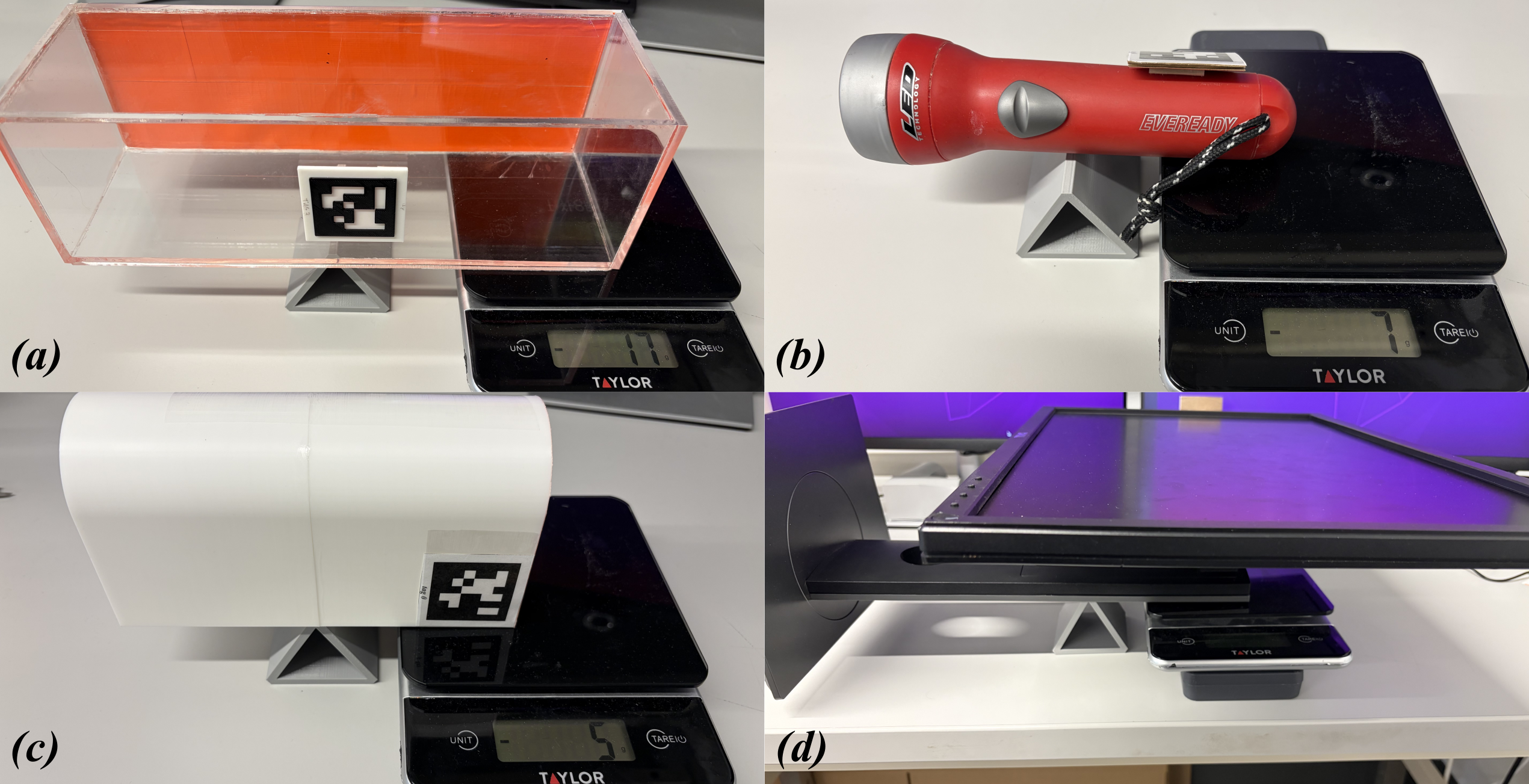}
    \caption{Validation of best CoM height estimates. Each object placed on small edge pivot at estimated CoM height, with weighing scale placed level with pivot. Lower measured weight indicates better CoM estimate. The measured results validate the computed results in the paper. }
    \label{fig: showcase}
\end{figure}

The attached video shows tipping processes in our experiments and validation results (Figure \ref{fig: showcase}). 

\section{Conclusions}
\label{sec:conclusions}
This work introduced a force-guided and vision-assisted active perception strategy for completing a three-dimensional CoM estimate by recovering CoM height and mass through controlled sub-critical tipping interactions. By exploiting the monotonic coupling between applied force and object angle during quasistatic tipping, the proposed method recovers CoM height and mass without requiring destructive toppling or multiple tipping configurations. Unlike prior approaches that rely on repeated near-marginal tipping or stable grasping, the presented framework enables parameter inference from a single controlled push while maintaining a configurable safety margin.

Experimental results across the tested objects demonstrate that sub-critical tipping provides sufficient observability for reliable estimation. Across the range of objects and safety margins tested, mass and CoM height estimates remained generally stable with low errors. These results highlight a fundamental tradeoff between safety and identifiability, suggesting that moderate safety margins can achieve sufficient accuracy without exposing objects to unstable equilibrium conditions. 

This work establishes sub-critical tipping as a practical and safe mechanism for 3D inertial parameter inference in non-prehensile manipulation.

However, the method can be further improved to address its current limitations. First, the failure cases revealed clear conditions under which tipping-based inference becomes less effective, such as curved or unstable object bases without sharp edges. Adapting the push direction could provide remedies. 

Cycle time per object could be improved by a higher or varied push speed while maintaining quasistatic object motion. 
The fit convergence time, could also be improved by increasing the signal-noise ratio in the force measurements.

A further refinement would be autonomous selection of informative push configurations, removing the reliance on AprilTags and the fixed-known pivot edge. Additional large-scale evaluation across diverse object sets will certainly be beneficial. Ultimately, we aim to develop a unified framework in which robots actively probe objects through safe physical interaction to infer inertial structure in real time, enabling more reliable manipulation in unstructured environments.


\ifCLASSOPTIONcaptionsoff
  \newpage
\fi


\bibliography{references.bib} 

\end{document}